\documentclass[letterpaper]{article} 
\usepackage{aaai24}  
\usepackage{times}  
\usepackage{helvet}  
\usepackage{courier}  
\usepackage[hyphens]{url}  
\usepackage{graphicx} 
\usepackage{natbib}  

\usepackage{caption} 
\usepackage{algorithm}
\usepackage{algorithmic}
\usepackage{makecell}
\usepackage{multirow}
\usepackage{booktabs}
\usepackage{amsmath,amsfonts,amsthm,bm}
\usepackage{xcolor}
\usepackage{newfloat}
\usepackage{listings}

\DeclareCaptionStyle{ruled}{labelfont=normalfont,labelsep=colon,strut=off} 
\floatstyle{ruled}
\newfloat{listing}{tb}{lst}{}
\floatname{listing}{Listing}
\title{Image Captioning with Multi-Context Synthetic Data}
\author{
    Feipeng Ma$^{1}$\thanks{This work was performed while Feipeng Ma was an intern with WeChat, Tencent Inc.} \quad Yizhou Zhou$^{2}$ \quad Fengyun Rao$^{2}$ \quad Yueyi Zhang$^{1,3}$\footnotemark[2] \quad Xiaoyan Sun$^{1,3}$\thanks{Corresponding authors: \{zhyuey, sunxiaoyan\}@ustc.edu.cn}
}
\affiliations{
    \textsuperscript{\rm 1}University of Science and Technology of China  \quad
    \textsuperscript{\rm 2}WeChat, Tencent Inc. \\
    \textsuperscript{\rm 3}Institute of Artificial Intelligence, Hefei Comprehensive National Science Center\\ 

    mafp@mail.ustc.edu.cn \quad \{harryizzhou, fengyunrao\}@tencent.com \quad
  \{zhyuey, sunxiaoyan\}@ustc.edu.cn \\
}

\usepackage{bibentry}

\begin{document}

\maketitle

\begin{abstract}
Image captioning requires numerous annotated image-text pairs, resulting in substantial annotation costs. Recently, large models (e.g. diffusion models and large language models) have excelled in producing high-quality images and text. This potential can be harnessed to create synthetic image-text pairs for training captioning models. Synthetic data can improve cost and time efficiency in data collection, allow for customization to specific domains, bootstrap generalization capability for zero-shot performance, and circumvent privacy concerns associated with real-world data. However, existing methods struggle to attain satisfactory performance solely through synthetic data. We identify the issue as generated images from simple descriptions mostly capture a solitary perspective with limited context, failing to align with the intricate scenes prevalent in real-world imagery. To tackle this, we present an innovative pipeline that introduces multi-context data generation. Beginning with an initial text corpus, our approach employs a large language model to extract multiple sentences portraying the same scene from diverse viewpoints. These sentences are then condensed into a single sentence with multiple contexts. Subsequently, we generate intricate images using the condensed captions through diffusion models. Our model is exclusively trained on synthetic image-text pairs crafted through this process. The effectiveness of our pipeline is validated through experimental results in both the in-domain and cross-domain settings, where it achieves state-of-the-art performance on well-known datasets such as MSCOCO, Flickr30k, and NoCaps.
\end{abstract}

\section{Introduction}

The realm of image captioning, which aims to craft informative textual descriptions for provided images, has witnessed remarkable progress. The crux of the challenge in image captioning hinges on comprehending the interplay between images and text, a dependency strongly rooted in the image-text pairs. Two approaches emerge to tackle this hurdle: one involves utilizing readily existing paired image-text data, while the other entails creating pairs from independent data sources.

There are two primary sources of existing paired image-text data: human-annotated and web-crawled.
Human-annotated data, employed in many studies~\cite{dai2017contrastive,anderson2018bottom}, can lead to significant improvements. However, the small size of available data limits its scalability and restricts domain generality.
Web-crawled data often suffer from low quality due to inaccurate correlations between images and text.
As a result, models~\cite{kang2022noise} trained on web-crawled data can exhibit limited performance in zero-shot settings. 
Moreover, large-scale crawling of image-text pairs possibly involves privacy and copyright issues.

In the absence of paired image-text data, unsupervised strategies often forge noisy image-text pairs from independent datasets for model bootstrapping~\cite{feng2019unsupervised} or domain alignment~\cite{laina2019towards}, leveraging a pretrained object detector. Moreover, \citet{meng2022object} suggest establishing object-text pairs by associating objects with sentences, rather than seeking candidates within the image collection. These techniques operate under the assumption that a pretrained detector can consistently discern visual concepts, thus establishing connections between disparate images and text. However, this assumption might not hold universally.

Inspired by \cite{he2022synthetic, zhao2022x}, which effectively employ diffusion model-driven synthetic data in image classification and segmentation, we have noticed the progress text-to-image models have achieved in crafting high-quality images from textual descriptions. This opens the door to the use of diffusion models for constructing image-text pairs in the image captioning domain. In comparison to human-labeled and web-crawled datasets, 
synthetic data offer efficiency in cost and time, enable customization for specific domains, bootstrap generalization capability for zero-shot performance, and sidestep privacy issues linked to real-world data. 
Customization for specific domains, referring to the in-domain ability, involves generating data tailored to specific domains, such as particular objects, attributes, or scenarios. Generalization capabilities, pertaining to cross-domain capability, entail generating synthetic data encompassing a broader range of scenarios, not limited to a specific objective.
However, there exists no prior work that specifically addresses the image captioning task solely using synthetic data generated via diffusion models.

\begin{figure*}[tbp]
    \vspace{-2mm}
    \centering
    \includegraphics[width=0.9\linewidth]{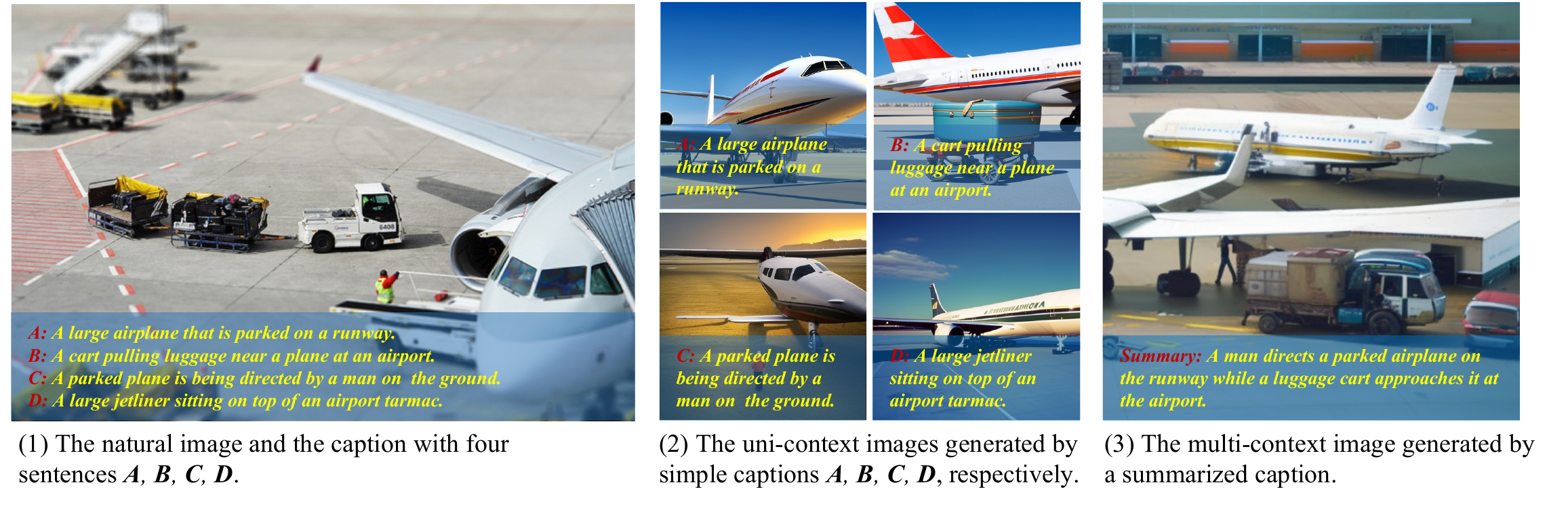}
    \caption{Examples of the natural image, the uni-context images, and the multi-context image.}
    \label{fig:intro}
\end{figure*}

The method to train an image captioning model using synthetic data involves two essential steps: (1) generating images along with captions through a diffusion model, utilizing established image captioning datasets. (2) subsequently training the image captioning model with the newly created synthetic dataset. However, a significant limitation arises when utilizing synthetic images, as they often lack the contextual depth necessary for ensuring precise image captioning accuracy. Our analysis highlights that synthetic images, originating from readily available basic captions, tend to exhibit constrained contexts, resulting in the omission of intricate and multifaceted scenes. These images are specifically referred to as ``uni-context" images. In contrast, natural images inherently encompass a multi-contextual essence, portraying a diverse array of objects, arrangements, interactions, and encapsulating complex and elaborate scenes. In this context, we provide a collection of examples for a comprehensive comparison between uni-context and multi-context images, depicted in Figure~\ref{fig:intro}. Notably, employing complex captions enables diffusion models to generate multi-context images, aligning with multi-faceted captions. Our focus for the image captioning task centers around the generation of multi-context captions to facilitate the synthesis of images with diverse contextual characteristics.

In this paper, we propose a pipeline for \textbf{I}mage \textbf{C}aptioning with Multi-Context \textbf{S}ynthetic \textbf{D}ata (\textbf{ICSD}).
Our pipeline starts with a text corpus containing accessible simple captions from diverse sources such as datasets, web crawls, and generated content. Comprising two key stages, the pipeline initiates with the generation stage and moves on to the training stage. \textbf{The generation stage} begins with obtaining complex captions. To optimally harness the corpus, we suggest selecting simple captions that might collectively depict the same scene rather than focusing on a small subset of direct complex captions. This process, termed selection and summarization, not only taps into the corpus' potential but also generates varied combinations of simple captions for diverse scenes. However, existing methods face challenges as they lack suitable metrics to determine if captions portray the same scene, and they need to be adaptable across different domains due to varied corpus sources. Leveraging the strengths of Large Language Models (LLMs) with their expansive knowledge and generalization abilities, we employ LLMs to execute selection and summarization tasks through provided instructions. Initially, we cluster captions based on text feature similarity, treating each caption as a query to construct LLM input candidates. The subsequent step instructs LLMs to pick captions from these clusters that could potentially form a complex scene. These chosen captions are then condensed into a single comprehensive caption. Subsequently, we generate multi-context images employing a generative model aided by these summarized captions. 
Moving into \textbf{the training stage}, our approach involves training models based on multi-context images derived from summarized sentences and the captions present in the corpus.
Each multi-context image is associated with its corresponding selected sentences, offering multiple related descriptions for every multi-context image. The training of our model relies solely on this synthetic data.

Our main contributions are summarized as follows:

\noindent (1) Pioneering the utilization of synthetic data in image captioning through the synergistic application of diffusion models and LLMs, introducing a novel approach in this field.

\noindent (2) Addressing the deficiency in complexity found in synthetic images generated from basic captions by analyzing the multi-context nature of natural images. We introduce a multi-context data generation pipeline tailored for enhancing image captioning. 

\noindent (3) Demonstrating the efficacy of our exclusively synthetic data-driven approach, we attain state-of-the-art performance in in-domain and cross-domain image captioning across three datasets: MSCOCO, Flickr30k and NoCaps. 

\section{Related Work}
\subsection{Supervised Image Captioning}
Conventional image captioning methods treat the task as a form of translation~\cite{vinyals2015show,karpathy2015deep}. These methods typically comprise a CNN-based encoder for image encoding and an RNN-based decoder for caption generation. Recent approaches~\cite{wang2022end,barraco2022unreasonable} adopt transformers architecture, yielding promising results. Additionally, research efforts integrate object~\cite{anderson2018bottom,song2021direction}, segmentation~\cite{wu2022difnet}, gaze patterns~\cite{alahmadi2022improve}, and attributes~\cite{fang2022injecting} to enhance image captioning models. Due to limited annotated data,  studies~\cite{li2022blip,hu2022scaling,wangsimvlm} pretrain models on expansive web-crawled datasets, followed by fine-tuning on smaller human-annotated datasets. Although these methods benefit from pretraining, their performance still heavily depends on the fine-tuning phase, which entails human-annotated data.

\subsection{Unsupervised Image Captioning}
Unsupervised image captioning seeks to train captioning models without the need for human-annotated data. Prior work utilizes independent image sources and text corpora for training, often leveraging object detectors to establish an initial link between the two modalities. \citet{feng2019unsupervised} pioneer this field by introducing policy gradient to reward generated captions aligned with correct visual concepts. Subsequently, \citet{laina2019towards} propose a shared multi-modal space constructed through visual concepts to align images and text. \citet{meng2022object} suggest harvesting objects corresponding to given sentences instead of finding candidate images. Nonetheless, these approaches depend heavily on object detectors, overlooking object attributes and relationships, constrained by detector generalization. Recent text-only training methods focus on training text decoder to reconstruct text from CLIP text encoder-derived features. During inference, they align image features extracted by CLIP's image encoder with text features in the same space. \citet{lidecap} introduce a training-free mechanism using training text features to project visual embeddings into text embedding space at inference. \citet{nukrai-etal-2022-text} and \citet{gu2023can} propose noise injection training to reduce the modality gap during inference. However, these methods rely on CLIP's cross-modality capacity and struggle to transfer to new domains without fine-tuning CLIP.

\subsection{Applications of Diffusion Models}

Diffusion models excel in generative capacities, spanning image creation, video synthesis, and text generation~\cite{ho2020denoising,dhariwal2021diffusion,villegas2022phenaki,chen2022analog}. Conditional versions enhance control and produce premium outcomes, extending their usefulness, as seen in text-to-image generation with models like DALL-E 2, Imagen, and Stable Diffusion~\cite{ramesh2022hierarchical,saharia2022photorealistic,rombach2022high}. Synthetic data from GLIDE demonstrated efficacy in image classification~\cite{nichol2022glide,he2022synthetic}, with further improvements achieved through ImageNet fine-tuning~\cite{azizi2023synthetic}. 
X-Paste~\cite{zhao2022x} leverages Stable Diffusion and CLIP to obtain synthetic images with accurate categories, which are transformed into instances for image segmentation.
These tasks need high-quality synthetic images but with less focus on matching meaning exactly. Only the object linked to a single label should appear in the image, without considering the whole scene. The text-to-image diffusion model manages this basic requirement. Unlike these tasks, image captioning requires intricate scenes in synthetic images that can be described from various perspectives.  
Because diffusion models cannot generate multi-context images from simple sentences, creating suitable training data for image captioning becomes quite a challenge.
\begin{figure*}[!htbp]
    \centering
    \includegraphics[width=0.8\textwidth]{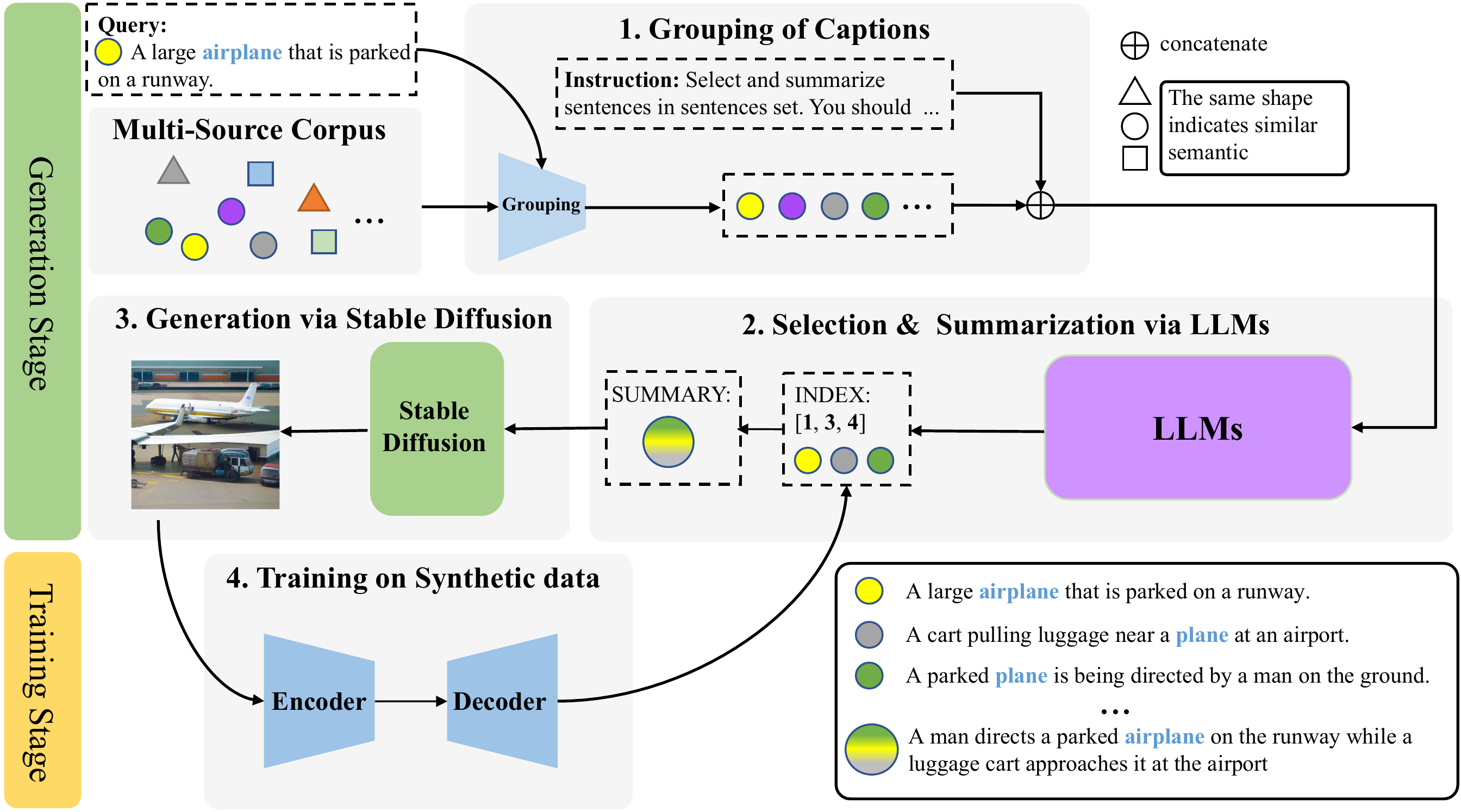}
    \caption{
        Overview of our proposed ICSD pipeline. The pipeline comprises two stages: the generation stage and the training stage. In the generation stage, we commence by performing the grouping of simple captions within the corpus. Next, LLMs are employed to select captions that depict the same scene from multiple perspectives, which are extracted from the obtained candidate sets. These selected captions are then condensed into a single sentence through summarization. These condensed sentences play a pivotal role in generating multi-context images using stable diffusion. Finally, in the training stage, we exclusively train the image captioning model on the synthetic multi-context image-text pairs. 
        }
    \label{fig:framework}
\end{figure*}

\section{Method}

\subsection{Overview}
Our pipeline, presented in Figure~\ref{fig:framework}, comprises two stages: the generation stage and the training stage. 
The generation stage begins with a given text corpus which comes from multiple sources for in-domain or cross-domain settings. For in-domain setting, we utilize human-annotated text to generate in-domain data. For cross-domain setting, we employ web-crawled text or text generated from LLMs with rich knowledge, to produce large-scale cross-domain data.
The generation stage consists of three steps: (1) Grouping of simple captions. In this step, for each simple caption acting as a query, we retrieve the most similar captions from the text corpus. These retrieved captions are then combined with the query caption to form a group. The captions in the same group possibly describe the same scene from diverse perspectives. 
(2) LLM-based selection and summarization. Providing the group of simple captions, which includes captions that potentially describe the same scene from diverse perspectives, we meticulously design prompt. 
These prompts guide LLMs to select simple captions that coherently align with a particular scene and summarize them into one sentence for image generation.
(3) Finally, we employ stable diffusion to generate images with the summarized captions. 
In the training stage, we train the image captioning model solely on the synthetic multi-context image-text pairs obtained from the generation stage.

\subsection{Generation Stage}
\noindent \textbf{Grouping of Simple Captions.} 
Given a text corpus  $T=\{t_1,t_2,...,t_N\}$ with $N$ captions, 
directly utilizing the whole text corpus as input is infeasible.
This is because the input context length limitation in LLMs prevents the use of the full corpus.
To address this issue, we propose to partition the text corpus into multiple groups of simple captions, with each group serving as a candidate set for selection and summarization.
Considering the large size of the text corpus, we form a group for each simple caption by retrieving instead of clustering algorithms.
Since captions describing the same scene often exhibit substantial semantic similarity with shared visual concepts, we employ CLIP to extract caption features and calculate the cosine similarity between each query caption and others within the corpus:
\begin{equation}
    s_{ij} = \frac{f(t_i)\cdot f(t_j)}{||f(t_i)|| \, ||f(t_j)||}
\end{equation}
where $s_{ij}$ is the cosine similarity between $f(t_i)$ and $f(t_j)$, $t_i$ is the query caption, $t_j$ is another caption in corpus, $f(\cdot)$ represents the text encoder of CLIP.
For query caption $t_i$, we retrieve the top $k$ similar sentences to form a group $G_i$ including $t_i$:
\begin{equation}
    G_i = \{t_i, t_m, \dots, t_n \}
\end{equation}
where the cardinality of $G_i$ is $k+1$, and $s_{ii} \ge s_{im} \ge ... \ge s_{in}$.
This results in $N$ groups, corresponding to the number of captions within the text corpus, with these groups containing overlapping captions.
We use a greedy algorithm to minimize redundancy and ensure that a small number of groups cover all corpus captions. The algorithm works as follows: (1) initialize a set $C$ identical to the corpus $T$. (2) repeatedly find and remove from $C$ the group $G_i$ with the most overlap with $C$, until $C$ is empty. Finally, we can find groups that can cover the entire corpus.

\noindent \textbf{Selection and Summarization via LLMs.} 
Selection and summarization pose significant challenges for existing technology. Selection aims to select simple captions that are able to describe the same image from various perspectives. The challenges of selection are (1) the demand for common sense: the selection process should incorporate the knowledge of natural scenes to decide what kind of objects should appear together in the scene and the given descriptions should not conflict. 
(2) the lack of metrics: the current metrics of text similarity are not designed for our target; the similarity metrics can not identify the descriptions that depict the same image.
Summarization aims to combine the selected simple captions into one complex caption for image generation. The challenge is that traditional text summarization approaches are ill-suited for our scenario since they aim to extract key information from long documents. And our summarization also demands strong generalization capabilities in open domains, since the corpus is so diverse.
Fortunately, the potency of LLMs empowers us to tackle these intricate challenges. LLMs are pretrained on a large scale of data, showing advancement in common knowledge and generalization ability. The instruction-following ability of LLMs makes it possible for us to formulate the selection and summarization task through language.
So we employ LLMs to tackle both tasks. We consider each group of captions as a candidate set for describing a specific image. We then formulate a prompt that enables us to accomplish both selection and summarization through LLMs. The prompt template is provided in Appendix C.

To avoid the hallucination problem of LLMs,  we incorporate the chain of thought technique~\cite{wei2022chain} into our design. We instruct the LLMs to first select sentences and subsequently summarize them into a single sentence. Additionally, we specifically prompt LLMs to provide the index of the chosen sentences, instead of generating these sentences anew, which may lead to hallucination problems.

\noindent \textbf{Image Generation with Stable Diffusion.}
In this crucial step, we harness the power of stable diffusion to generate synthetic images based on the summarized sentences derived from the text corpus. 
We refrain from using any prompt engineering on the captions that are inputted to stable diffusion, aiming to minimize the influence of human intervention during large-scale generation. Consequently, we can acquire a substantial volume of multi-context images.

\subsection{Training Stage}
\noindent \textbf{Architecture and data.} 
We follow BLIP~\cite{li2022blip} to adopt the encoder-decoder architecture~\cite{vaswani2017attention}. 
The encoder is initialized from the weights of ViT-B/32, while the decoder is initialized from the weights of BERT-base~\cite{devlin2018bert}.
We exclusively utilize synthetic data for training, specifically focusing on multi-context image-text pairs. 

\begin{table*}[!ht]
    \vspace{-1mm}
    \vspace{-2mm}
    \begin{center}
    \resizebox{0.9\textwidth}{!}{
    \small
        \begin{tabular}{l cc cccccccc}
    \toprule
    \multirow{2}{*}{Methods}& \multicolumn{2}{c}{Data} & \multicolumn{4}{c}{MSCOCO}&\multicolumn{4}{c}{Flickr30k}\\
     &I.&T. & B@4 & M & R & C & B@4 & M & R & C \\
    \midrule
    \citet{feng2019unsupervised}  &\checkmark &\checkmark & 18.6 & 17.9 &43.1 & 54.9  & - & - & - & - \\
    \citet{laina2019towards}  &\checkmark &\checkmark & 19.3 & 20.2 & 45.0 & 61.8  & - & - & - & - \\
    ESPER-Style~\cite{yu2022multimodal}&\checkmark &\checkmark & 21.9&21.9&- &78.2  & - & - & - & - \\
    ZeroCap~\cite{tewel2022zerocap}  & &\checkmark &7.0 & 15.4& 31.8 &34.5 &5.4&11.8&27.3&16.8\\
    Magic~\cite{su2022language}  & & \checkmark& 12.9 & 17.4 & 39.9 & 49.3&  6.4 &13.1&31.6&20.4 \\
    CLIPRe~\cite{su2022language}   & &\checkmark & 4.9 & 11.4&29.0&13.6 & 5.2 & 11.6 &27.6&10.0 \\
    CapDec~\cite{nukrai-etal-2022-text}  & & \checkmark& \underline{26.4} &25.1&\underline{51.8} &91.8 & 17.7 & 20.0&43.9&39.1\\
    DeCap*~\cite{lidecap}  & & \checkmark& 25.3 &\underline{25.2}&51.1 &92.9 &\underline{20.0} & \textbf{21.7}&\underline{46.2}&\underline{49.3}\\
    CLOSE~\cite{gu2023can}  & & \checkmark & - & - & - & \underline{95.3} & - & - & - & -\\
    \midrule
    ICSD  & & \checkmark &\textbf{29.9} &\textbf{25.4}&\textbf{52.7} &\textbf{96.6} & \textbf{25.2} & \underline{20.6} & \textbf{46.7} & \textbf{54.3} \\
    \bottomrule
    \end{tabular}
    }
    \end{center}
    \vskip -0.1in
    
    \caption[caption]{The results of in-domain image captioning on MSCOCO and Flickr30k. ``I.'' and ``T.'' denote the need for external image data and text data, respectively. 
    ``*'' means results reproduced using the provided code. B@4: BLEU@4; M: METEOR; R: ROUGE; C: CIDEr.
    Numbers in \textbf{bold} and \underline{underlined} text represent the best and second-best results, respectively.
     }
    \label{tab:in-domain captioning}
    \end{table*}

\noindent \textbf{Objective.} We utilize synthetic data to train our model using Cross Entropy Loss:
\begin{equation}
    \mathcal{L} = -  \sum_{i=1}^{n} \log(P(y_{i}|y_{1:i-1},v))
\end{equation}
where $P$ denotes the probability distribution from the language decoder, $y_i$ denotes the ground-truth word at time step $i$, $y_{1:i-1}$ refers to the prior words, $n$ stands for the length of the ground-truth sentence, and $v$ is the synthetic image.

\section{Experiments}
We conduct experiments in two settings: in-domain and cross-domain image captioning. In the in-domain setting, the training and test data are derived from the same dataset. In the cross-domain setting, the training and test data are sampled from different datasets, requiring the model to effectively generalize across diverse data sources.

\subsection{Settings}
\noindent \textbf{Datasets.} 
For in-domain image captioning, we utilize MSCOCO~\cite{lin2014microsoft} and Flickr30k~\cite{young2014image} datasets. MSCOCO contains 123,287 images, each annotated with five captions. Following~\cite{karpathy2015deep}, we split MSCOCO into 118,287 for training, 4,000 for validation, and 1,000 for testing. Flickr30k contains 31,783 images, with each image accompanied by five captions.
Regarding cross-domain image captioning, we train our model on SS1M~\cite{feng2019unsupervised} dataset and evaluate its performance using MSCOCO and NoCaps~\cite{agrawal2019nocaps}. SS1M is a web-scraped text corpus and contains 2,322,628 image descriptions from Shutterstock using eighty object class names in MSCOCO as keywords. In line with DeCap~\cite{lidecap}, we exclude sentences containing more than fifteen words on SS1M. 
We use the validation set of NoCaps to evaluate performance in three settings: in-domain, near-domain, and out-of-domain.
For all datasets, we solely utilize their text during training and do not acquire any natural images.

To measure the quality of generated captions, we follow prior studies~\cite{dai2017towards,meng2022object} and employ metrics such as BLEU~\cite{papineni2002bleu}, METEOR~\cite{banerjee2005meteor}, ROUGE~\cite{lin2004rouge}, and CIDEr-D~\cite{vedantam2015cider}.

\begin{table*}[tbp]
    \begin{center}
    \resizebox{0.8\textwidth}{!}{%
    \renewcommand{\arraystretch}{1.0}
    \begin{tabular}{l c cccc cccc}
    \toprule
    \multirow{2}{*}{Methods}& \multirow{2}{*}{Dataset} & \multicolumn{4}{c}{MSCOCO} & \multicolumn{4}{c}{NoCaps val (CIDEr)}\\
     &  &B@4&M&R&C &In&Near&Out&Overall \\ 
    \midrule
    ZeroCap~\cite{tewel2022zerocap} &-  & 2.6 & 11.5& - & 14.6  &-&-&-&-\\
    ConZIC~\cite{zeng2023conzic} & -  &1.3 &11.5&- & 12.8     &-&-&-&-\\
    CLIPRe~\cite{su2022language}  &CC3M-text &4.6&13.3&-&25.6 &23.3&26.8&36.5&28.2\\
    DeCap~\cite{lidecap}& CC3M-text&8.8&16.0&-&42.1    &34.8&37.7&\textbf{49.9}&39.7\\
    DeCap~\cite{lidecap}& SS1M &\underline{8.9}&\underline{17.5}&-&\underline{50.6}        &\underline{41.9}&\underline{41.7}&\underline{46.2}&\textbf{42.7}\\
    \midrule
    ICSD & SS1M &\textbf{13.6} &\textbf{18.3}&\textbf{38.0}&\textbf{54.2} &\textbf{42.9}&\textbf{44.3}&35.6&\textbf{42.7}\\
    \bottomrule
    \end{tabular}
    }
    \caption[caption]{The results of cross-domain image captioning on MSCOCO and NoCaps. 
    } 
    \label{tab:ss1m}
    \end{center}
    \vskip -0.1in
\end{table*}

\noindent \textbf{Implementation Details.}
During the generation stage, we use different group sizes for various datasets: 30 for MSCOCO, 20 for Flickr30k, and 10 for SS1M. We employ the GPT-3.5-turbo model for selecting and summarizing captions via API access. For image generation, we utilize Stable Diffusion v1.4 at a 512$\times$512 resolution with 20 sampling steps, and we speed up the diffusion model's sampling process using DPM-Solver~\cite{ludpm}.
We train the model for 30 epochs using Adam~\cite{kingma2014adam} and a batch size of 36. The learning rate is 1e-5, and a warm-up strategy is applied during training. 
Additionally, the input synthetic images are resized to 384$\times$384. For inference, we follow the BLIP to use beam search with a beam size of 3.
All experiments are conducted using eight NVIDIA A100 GPUs.

\subsection{Comparisons with State-of-the-art Models}
\noindent \textbf{In-domain Image Captioning.}
We perform in-domain image captioning on MSCOCO and Flickr30k datasets, comparing our ICSD with state-of-the-art unsupervised methods:
\citet{feng2019unsupervised} and \citet{laina2019towards} train models on independent image and text data, using visual concepts to establish connections between images and text. ZeroCap~\cite{tewel2022zerocap}, Magic~\cite{su2022language}, and ESPER-Style~\cite{yu2022multimodal} incorporate GPT-2~\cite{radford2019language} as the language decoder. CLIPRe~\cite{su2022language} is a CLIP-based method for retrieving captions. CapDec~\cite{nukrai-etal-2022-text}, DeCap~\cite{lidecap} and CLOSE~\cite{gu2023can} conduct text-only training, leveraging the powerful cross-modal capability of CLIP.

The comparison results for MSCOCO and Flickr30k datasets are presented in Table~\ref{tab:in-domain captioning}. We generate 150,000 multi-context images for MSCOCO and 140,000 images for Flickr30k, with each synthetic image paired with 5 to 10 captions. Our method significantly outperforms other unsupervised approaches across the majority of metrics. In the B@4 metric, our ICSD surpasses previous state-of-the-art methods by 13.3\% on MSCOCO and 26.0\% on Flickr30k.

\noindent \textbf{Cross-domain Image Captioning.}
The captions of MSCOCO and Flickr30k can naturally be grouped, as each image has at least five descriptions in these human-annotated datasets. To evaluate the effectiveness of our ICSD, we train the model on the web-crawled SS1M captions, which are not inherently grouped, and perform cross-domain image captioning on MSCOCO and NoCaps. 
We create 150,000 multi-context images for SS1M. Due to SS1M's large scale and the API call limitations of GPT-3.5-turbo, we additionally generate uni-context images for each caption.
We compare our method with several other approaches in this experimental setting: (1) ZeroCap and ConZIC~\cite{zeng2023conzic} directly use pretrained vision-language models without fine-tuning; (2) CLIPRe and DeCap are trained on the large CC3M-text corpus~\cite{changpinyo2021conceptual}; (3) DeCap and \citet{feng2019unsupervised} also employ SS1M to train the models.

The results in Table~\ref{tab:ss1m} demonstrate the effectiveness of our method.
When evaluating ICSD on MSCOCO, our method achieves obvious improvements across all metrics. Especially on BLEU and CIDEr metrics, improved from 8.9 to 13.6 and 50.6 to 54.2, respectively. 
This implies that the effectiveness of our method is not limited to in-domain image captioning, it remains efficient even when applied to a wide range of data collected from the web.
In the NoCaps evaluation, our method performs well on in-domain and near-domain sets but not as strongly on the out-domain set. This discrepancy arises because SS1M is collected based on the objects of MSCOCO, which is not strictly designed for cross-domain setting, affecting its performance on the out-domain set of NoCaps. However, our method is able to address this limitation by utilizing LLMs to generate a corpus containing diverse objects for generalization and applying our pipeline with this corpus. We report the results of this experiment in Appendix A.
\begin{figure*}[tbp]
    \centering
    \includegraphics[width=0.75\linewidth]{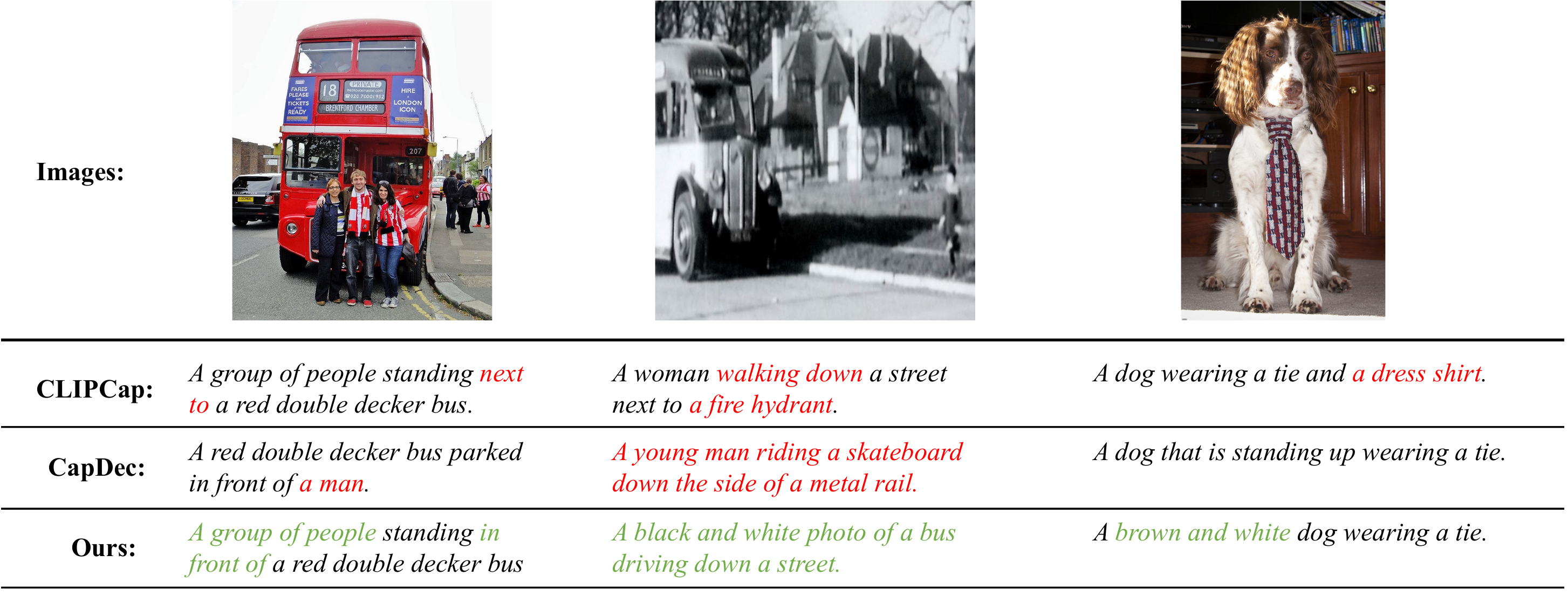}
    \caption{
    Comparisons of captions generated by CLIPCap~\cite{mokady2021clipcap}, CapDec~\cite{nukrai-etal-2022-text}, and our proposed ICSD, using exemplary images from MSCOCO dataset.
    }
    \label{fig:vis}
\end{figure*}

\subsection{Ablation Study}

\noindent \textbf{The effect of components of ICSD.}
Table~\ref{tab:ablation_module} presents the results of an ablation study evaluating each component on MSCOCO. The \textit{Baseline} indicates that we train the model using only uni-context image-text pairs. 
We then investigate three variations of selection and summarization: (1) selection without summarization (\textit{Sel. w.o. Sum.}), where we employ LLMs to select captions for training but do not merge them into a single sentence for multi-context image generation. 
The captions selected from the same group are associated with the uni-context image, which is generated with query caption of the group. 
(2) summarization without selection (\textit{Sum. w.o. Sel.}), in which LLMs directly summarize the top five similar captions in the group, potentially leading to errors since the most similar captions are not necessarily conflict-free. (3) both selection and summarization (\textit{Sel. \& Sum.}), representing our ICSD approach. 
To further emphasize the importance of selection in collecting a high-quality caption set for summarization, we conduct experiments using the ground-truth group (\textit{w. GTG}) as a substitute for selection. MSCOCO is a human-annotated dataset where each image has at least five captions, inherently containing captions written for the same image by different annotators. In the \textit{w. GTG} experiment, we use these captions, written for the same image, for summarization.

The results of \textit{Baseline} exhibit limited performance across most metrics. In comparison with previous state-of-the-art methods, the Meteor and CIDEr scores are lower than DeCap by 0.5 and 2.1, respectively, while the ROUGE-L score is lower than CapDec by 0.4. The limited performance indicates that the uni-context images, generated by single captions, are insufficient for image captioning.

The \textit{Sel. w.o. Sum.} approach results in improved performance, particularly in the BLEU and CIDEr metrics. This improvement can be attributed to the fact that \textit{Sel. w.o. Sum.} employs LLMs to select captions that may describe the same scene and associate them with the image generated using the query caption. This process constructs pseudo multi-context pairs in which the uni-context image is associated with multiple captions, albeit with less accurate correlations.

The \textit{Sum. w.o. Sel.} presents another type of pseudo multi-context image-text pair. By directly summarizing similar captions into a single caption through LLMs and generating multi-context images, this approach may introduce errors, as the similarity of captions cannot guarantee that they are conflict-free when describing the same scene. Consequently, the summarized caption may not be fully compatible with the original simple captions. However, the generated images are multi-context images, resulting in improved performance compared to \textit{Sel. w.o. Sum.} method. This observation demonstrates the importance of multi-context images for image captioning, but the inaccurate correlations lead to inferior performance in this approach.

The performance of our proposed ICSD approach, which incorporates both \textit{Sel. \& Sum.}, is significantly improved. As the selection process can gather compatible captions for summarization, which is absent in the \textit{Sum. w.o. Sel.} method, summarization can create multi-context captions for multi-context image generation, an aspect that is lacking in the \textit{Sel. w.o. Sum.} approach. 
The results indicate that by combining selection and summarization, our ICSD can generate multi-context data that are highly beneficial for image captioning. To further explore our method, we use the ground-truth grouping of MSCOCO and summarize captions. 
This strategy notably enhances performance across all metrics, underscoring the importance of effective grouping.

\begin{table}[tbp]
    \centering
    \renewcommand{\arraystretch}{1.0}
    \resizebox{0.4\textwidth}{!}{
    \begin{tabular}{l cccc}
        \toprule
        Method&B@4&M&R&C \\ 
        \midrule
        Baseline  & 26.9 &24.7 &51.4 &90.8\\
        \midrule
        Sel. w.o. Sum. & 29.5 & 24.7	& 52.3 & 94.2 \\
        Sum. w.o. Sel. & 29.6 & 25.0 & 52.9 & 95.8 \\
        Sel. \&  Sum. (ICSD)    & 29.9 &25.4	&52.7 &96.6\\
        \midrule
        ICSD \textit{w. GTG}  &\textbf{30.2} & \textbf{25.7} &\textbf{53.0} &\textbf{97.3}\\
        \bottomrule
    \end{tabular}
    }
    \caption{The effect of components of ICSD on MSCOCO.
    }
    \label{tab:ablation_module}
\end{table}

\noindent \textbf{The impact of the number of multi-context images.}
We conduct experiments using different numbers of multi-context images for training and evaluate the model on the test split of MSCOCO. Table~\ref{tab:numbers} presents the results, where we increase the number of multi-context images from 10,000 to 200,000.
The results demonstrate that incorporating more multi-context images during training can improve the performance of in-domain image captioning. Particularly, we observe significant gains in the B@4 and CIDEr metrics when increasing the number of multi-context images from 10,000 to 50,000. The best performance achieved thus far is obtained when using 150,000 multi-context images. 
However, expanding the number of multi-context images to 200,000 yields only marginal improvements, primarily due to the constrained diversity of the text corpus.

\noindent \textbf{Visualization.}
Our model, trained on synthetic data, is capable of generating accurate captions for natural images. In Figure~\ref{fig:vis}, we present examples of generated captions on the test split of MSCOCO, comparing our approach with CLIPCap and CapDec. The incorrect portions of the captions are highlighted in red, while the improvements made by our method are emphasized in green.
For the first image, our method accurately describes the location and the number of people. Regarding the second and third images, our approach outperforms the others by capturing more detailed descriptions, such as the colors ``black and white" and ``brown and white".

In Appendix B, we further explore a range of alternative methods, including retrieving images instead of generating them, as well as creating a detailed caption from a given input caption, as opposed to merging multiple simple captions.

\begin{table}[tbp]
    \centering
    \fontsize{7pt}{7pt}\selectfont
    \renewcommand{\arraystretch}{0.8}
    \resizebox{0.35\textwidth}{!}{
    \begin{tabular}{r cccc}
        \toprule
        Number &B@4&M&R&C \\ 
        \midrule
        10,000 & 28.0 & 23.8 & 51.1 & 87.9  \\
        50,000 & 29.3 &24.7&52.3&93.5 \\
        100,000 & 29.4 &25.0&52.3&95.1 \\
        150,000 & 29.9 &\textbf{25.4}&\textbf{52.7}&\textbf{96.6} \\
        200,000 & \textbf{30.1} &25.1&\textbf{52.7}&96.5 \\
        \bottomrule
    \end{tabular}
    }
    \caption{The impact of the number of multi-context images.}
    \label{tab:numbers}
\end{table}

\section{Conclusion}
We observe that synthetic images generated from single captions lack the ability to be described from multiple perspectives, unlike real-world images. To address this issue, we propose a pipeline called ICSD that generates multi-context training data by combining LLMs and diffusion models for image captioning. The pipeline has two stages: generation and training. In the generation stage, we group captions in the corpus and select diverse perspectives using LLMs. These perspectives are summarized into a single sentence, which is then used to generate multi-context images through diffusion models. This results in high-quality synthetic multi-context image-text pairs where each image can be described from various perspectives. In the training stage, we train image captioning models using the synthetic data generated in the generation stage. Extensive experiments on in-domain and cross-domain image captioning demonstrate the effectiveness of our ICSD pipeline.

\section{Acknowledgments}
This work was in part supported by the National Natural Science Foundation of China under grants 62032006 and 62021001.

\bibliography{aaai24}

\newpage

\begin{table*}[tbp]
    \begin{center}
    \resizebox{0.8\textwidth}{!}{%
    \renewcommand{\arraystretch}{1.0}
    \begin{tabular}{l c cccc cccc}
    \toprule
    \multirow{2}{*}{Methods}& \multirow{2}{*}{Dataset} & \multicolumn{4}{c}{MSCOCO} & \multicolumn{4}{c}{NoCaps val (CIDEr)}\\
     &  &B@4&M&R&C &In&Near&Out&Overall \\ 
    \midrule
    ZeroCap~\cite{tewel2022zerocap} &-  & 2.6 & 11.5& - & 14.6  &-&-&-&-\\
    ConZIC~\cite{zeng2023conzic} & -  &1.3 &11.5&- & 12.8     &-&-&-&-\\
    CLIPRe~\cite{su2022language}  &CC3M-text &4.6&13.3&-&25.6 &23.3&26.8&36.5&28.2\\
    DeCap~\cite{lidecap}& CC3M-text&8.8&16.0&-&42.1    &34.8&37.7&\textbf{49.9}&39.7\\
    DeCap~\cite{lidecap}& SS1M &8.9&17.5&-&50.6&41.9&41.7&46.2&42.7\\
    \midrule
    ICSD & SS1M &13.6 &\underline{18.3}&38.0&\underline{54.2} &42.9&44.3&35.6&42.7\\
    ICSD & GenT &\underline{15.0}&18.2&\textbf{41.4}&47.2 &\underline{51.3} &\underline{47.0} &40.3 &\underline{46.4}\\
    ICSD & SS1M+GenT &\textbf{16.0} &\textbf{18.9}&\underline{40.9}&\textbf{58.3} &\textbf{54.3}&\textbf{54.0}&\underline{48.3}&\textbf{53.5}\\
    \bottomrule
    \end{tabular}
    }
    \caption[caption]{Cross-domain image captioning on MSCOCO Karpathy-test split and NoCaps validation set. 
    ``I.'' and ``T.'' denote image data and text data, respectively. 
    B@4: BLEU@4; M: METEOR; R: ROUGE; C: CIDEr.
    Numbers in \textbf{bold} and \underline{underlined} text represent the best and second-best results, respectively.
    } 
    \label{tab:appendix_ss1m}
    \end{center}
    \vskip -0.1in
\end{table*}

\section{A \quad Cross-Domain Image Captioning}
We evaluate our method on cross-domain image captioning to assess its generalization capability. As shown in Table~\ref{tab:appendix_ss1m}, our ICSD, trained on the text of SS1M, performs well on in-domain and near-domain sets of NoCaps, but its performance on the out-domain set is limited. We identified the core issue as the SS1M dataset being collected based on the objects of MSCOCO, which is not designed for cross-domain settings, thus affecting its performance on the out-domain set of NoCaps.
Since our ICSD can be applied to various text corpora, including web-crawled text and text generated from LLMs, we can address this issue by employing LLMs to generate diverse text, thereby enhancing the generalization capability.

\subsection{Generate Text Corpus with LLMs}
To achieve cross-domain image captioning, a large and diverse text corpus is essential. Our pipeline can effectively utilize LLMs to obtain such a text corpus.

\noindent \textbf{Object Collections.} 
We first construct a set of objects for generating a text corpus. Unlike SS1M, which is based on the 80 objects from the MSCOCO dataset, our aim is to collect a large-scale array of objects that are not limited to a specific domain. Our objects are sourced from existing datasets or generated using LLMs. Specifically, we utilize the 80 objects from MSCOCO and randomly collect 2,500 objects from the Visual Genome~\cite{krishna2017visual}. Additionally, we employ LLMs to generate 400 common objects. As a result, we gather nearly 3,000 objects for text corpus generation. 
These objects are not collected for a specific domain but rather serve a more general purpose in cross-domain image captioning.

\noindent \textbf{Text Corpus Generation.} 
We use the collected objects as the foundation for corpus generation. We instruct LLMs to generate high-quality text based on the given objects. To achieve better results, we randomly select 80 objects as context for LLMs, rather than providing all objects at once.
The prompt is:

\noindent \textit{
Given 100 objects [ person, man, woman, people, ...] 
    Using the provided objects to generate captions that describe real-world images in an objective manner.
    The generated sentences are captions for images, the scene described in the sentence must be reasonably realistic.
    The generate 100 sentences should follow these requirements: }

\noindent \textit{1. Descriptive: The captions provide clear and accurate descriptions of the objects and actions in the scene.} 
    
\noindent \textit{2. Concise: The captions are no longer than 15 words, but more than 8 words. }
    
\noindent \textit{3. Objective: The descriptions focus on the factual aspects of the scene, avoiding subjective interpretations or emotions.}
    
\noindent \textit{4. Present tense: The captions describe events happening in the present moment, not past or future events.}
    
\noindent \textit{5. No adverbs: The sentences do not contain adverbs, making the descriptions more straightforward.}
    
\noindent \textit{6. Avoid starting with certain phrases: The captions do not begin with phrases like `There', `An image', or `A photo of'.}
    
\noindent \textit{Please output the sentences with `;' as the separator.}

\noindent \textbf{Post-processing.}
We then apply post-processing to ensure the quality of the generated text, removing sentences that are longer than 15 words or shorter than 8 words. Ultimately, we construct a \textbf{Gen}erated \textbf{T}ext corpus containing 450,000 sentences, which we refer to as \textbf{GenT}.

\subsection{Experimental Results}

In Table~\ref{tab:appendix_ss1m}, we apply our pipeline to GenT to develop an image captioning model with enhanced generalization capability. We create 50,000 multi-context images with 3 to 8 captions for each image We observe significant performance improvements on NoCaps, particularly the CIDEr score on the out-domain set, which increases from 35.6 to 40.3. Since GenT is generated from a large number of objects by LLMs, our pipeline effectively leverages GenT, resulting in robust generalization capability for cross-domain image captioning.

Additionally, we combine SS1M and GenT for our method, leading to substantial performance improvements on both MSCOCO and NoCaps. On MSCOCO, the BLEU@4 score increases from 13.6 to 16.0, and the CIDEr score rises from 54.2 to 58.3. On NoCaps, the CIDEr scores across all three domains improve significantly, with increases of over 10 points. Notably, the CIDEr score on the out-domain set increases from 35.6 to 48.3, demonstrating a strong generalization capability.

\section{B \quad More Analysis of the Proposed ICSD}
\noindent\textbf{Can retrieving images from a corpus be an alternative to generating images with diffusion models?} A major concern is whether obtaining synthetic images by diffusion models is necessary. An alternative way is to use condensed captions to retrieve corresponding images from an accessible image corpus, which requires less GPU memory and time. In Table~\ref{tab:sup_gap}, we implement the retrieval-based method by sampling 1 million images from laion-2b~\cite{schuhmann2022laion} and using CLIP to retrieve images for condensed captions. 
The results of the retrieval-based method are inferior to our ICSD. 
The reason is that accurately retrieving images that match detailed descriptions is difficult.
For example, CLIP suffers from aligning images with detailed text~\cite{yao2021filip}.

\noindent\textbf{Can generating detailed caption from an input caption replace merging simple captions by LLMs?}
Asking LLMs to make up a more detailed scene description that matches an input caption is a simpler method. In Table~\ref{tab:sup_gap}, we conduct experiments of this generative-based method on MSCOCO, generate a condensed caption for each simple caption, and generate images from condensed captions for training. The generative-based method achieves 28.9 and 94.7 on B@4 and CIDEr, lower than our ICSD. However, these methods are not in conflict, they serve different scenarios. Our ICSD focuses on in-domain settings, aligning images with the target text corpus distribution, while the generative-based method emphasizes generalization without considering specific text distribution.

\section{C \quad Prompt for Selection and Summarization}
In this section, we provide the prompt template used for selection and summarization:

\noindent \textit{Select and summary sentences in the given sentences set. You should find 3 to 8 sentences that form a description from the same or different views of the same image. The meanings of the selected sentences in a group should not conflict with each other. Summarize the sentences as one not exceed 50 words to describe the scene in objective style. The summary sentence must be objective and concise, without ambiguity and uncertainty. Return the selected index and the summarized sentence in json format like {`index': list,`summary': str}. Return directly the json format results without explanation. The given sentences set are: 1. ... 2. ...}

ndidate set and instruct LLMs to perform selection and summarization. To avoid the hallucination problem, we guide LLMs to first select sentences and then summarize them into a single sentence. Furthermore, we ask LLMs to output the index of the selected sentences, rather than generating these sentences anew.

\begin{table}[h]
    \centering
    \renewcommand{\arraystretch}{1.0}
    \resizebox{0.46\textwidth}{!}{
    \begin{tabular}{l cccc}
        \toprule
        Method&B@4&M&R&C \\
        \hline
        Retrieval-based method  & 27.3 &23.4 &50.4 &85.7\\
        Generative-based method  & 28.9 &25.1 &52.2 &94.7\\
        ICSD(Ours) & \textbf{29.9} & \textbf{25.4}	&\textbf{52.7} &\textbf{96.6}\\
        \bottomrule
    \end{tabular}
    }
    \caption{Comparison with retrieval-based and generative-based methods.
    }
    \label{tab:sup_gap}
\end{table}

\section{D \quad Visualization}
\begin{figure*}[htbp]
    \centering
    \includegraphics[width=0.9\linewidth]{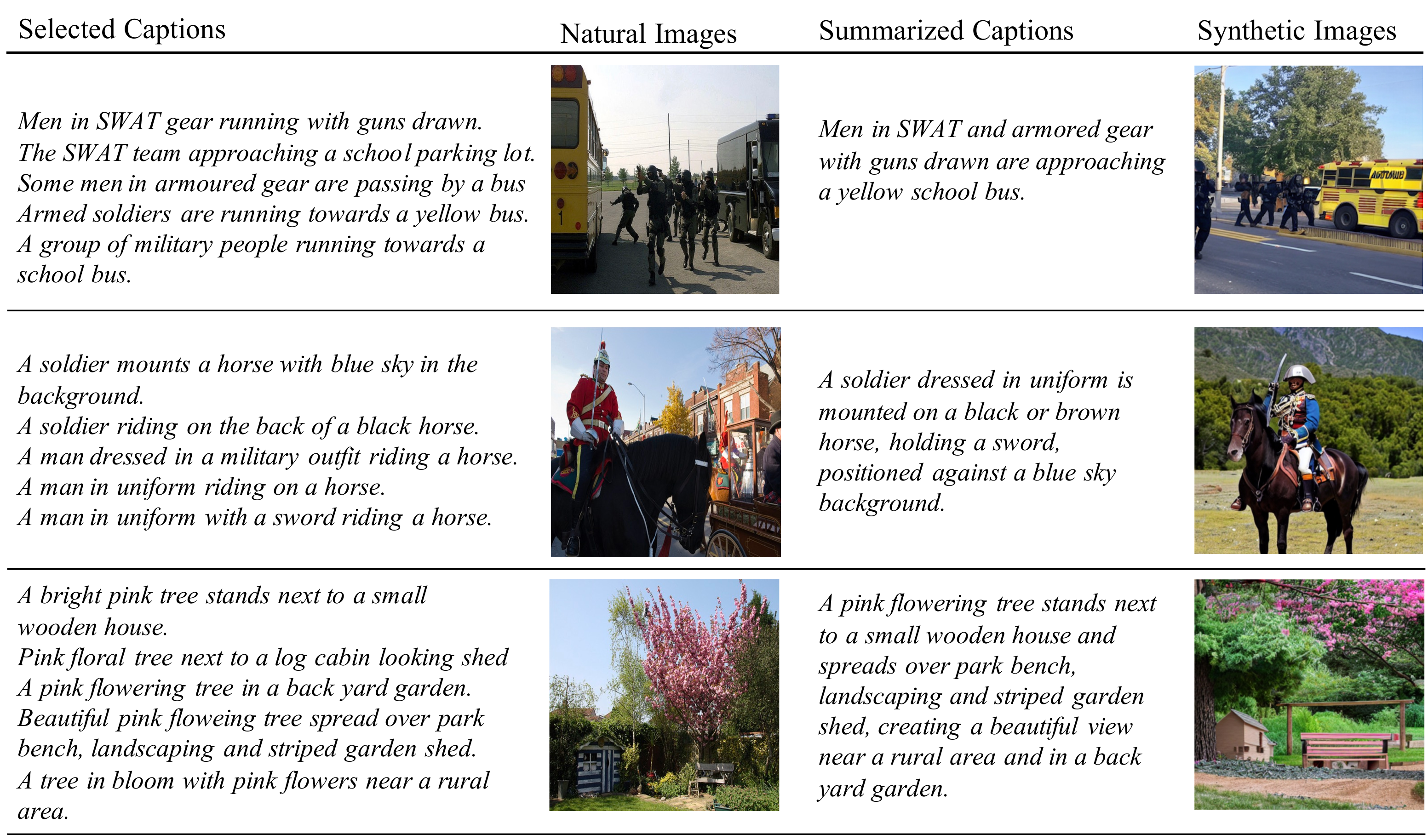}
    \caption{Visualization of selection and summarization with the corresponding natural images and synthetic images.}
    \label{fig:supp_vis}
\end{figure*}

We present some cases of selection and summarization in Figure~\ref{fig:supp_vis}, with each row representing a specific instance. 
The columns in the figure are: (1) Selected Captions: captions selected from the initial grouping by LLMs. (2) Natural Images: The real images corresponding to a specific selected caption. (3) Summarized Captions: The summarized captions of selected captions by LLMs. (4) Synthetic Images: The synthetic images generated with summarized captions through Stable Diffusion.
In selected captions, multiple captions correspond to the same natural image, and we choose this natural image to represent the ground truth scene. 
Overall, we observe that the synthetic images are very close to the corresponding natural images in terms of scene and can be described by multiple selected captions. 
This observation verifies our hypothesis that by combining LLMs and Stable Diffusion, we are able to obtain synthetic images that can be described from multiple perspectives, and such synthetic images are closer to natural images.




\end{document}